\documentclass[12pt,a4paper]{article}
\usepackage[vietnamese,english]{babel}
\usepackage[T5,T1]{fontenc}

\usepackage{amsmath, amssymb, fullpage, graphics}
\usepackage{caption, graphicx, fancyhdr, fancybox}
\usepackage{color}
\usepackage{tabularx}
\usepackage{booktabs}
\usepackage{multicol}
\usepackage{multirow}
\usepackage{makecell}
\usepackage{pifont}
\usepackage[ruled,vlined,linesnumbered]{algorithm2e}
\usepackage{listings}
\usepackage{enumitem}
\usepackage{longtable}
\usepackage[para]{footmisc} 
\usepackage{hyperref}

\usepackage{pgf}
\usepackage[numbers]{natbib}

\makeatletter
\fancypagestyle{footmark}{
  \ps@@fancy 
  \fancyfoot[C]{\footmark}
}

\makeatother
\begin{document}

\begin{center}
{\large\sf VietAIDetector: An Open-Source Zero-Shot Detector for Vietnamese AI-Generated Text}
\vspace{2mm}

Trieu Hai Nguyen$^{\text{a},\dagger}$, Van-Dung Hoang$^{\text{b},*}$ \\[2mm]

${^\text{a}}$ \textit{Nha Trang University, 02 Nguyen Dinh Chieu Street, North Nha Trang 57134, Vietnam} \\
${^\text{b}}$ \textit{Ho Chi Minh City University of Technology and Engineering, 01 Vo Van Ngan Street, Thu Duc, Ho Chi Minh City, Vietnam} \\
[2mm] 
e-mails: $^\dagger$\textit{trieunh@ntu.edu.vn}; $^*$\textit{dunghv@hcmute.edu.vn}

\end{center}
\begin{abstract}
Distinguishing AI-generated content from human-written text in Vietnamese remains difficult due to the limited availability of dedicated open-source tools. In this paper, we present VietAIDetector, an open-source software tool designed for practical Vietnamese AI-generated text detection. Operating on a zero-shot scoring engine derived from VietBinoculars~\cite{nguyen2025vietbinoculars}, the software requires no domain-specific retraining and accepts raw text, DOCX files, native PDF documents, and scanned PDFs via integrated optical character recognition. To handle long documents beyond conventional input lengths, the tool combines sliding-window chunking with document-level score aggregation. VietAIDetector offers an interactive Gradio web interface with configurable operating thresholds, including Youden's J statistic, the Closest Point approach, and a low false-positive rate criterion, with results exportable as downloadable PDF reports. Empirical validations confirm that this chunking pipeline mitigates threshold drift on multi-thousand-token documents while maintaining practical runtime throughput under GPU deployment. The tool is publicly available at \url{https://github.com/trieuntu/VietAIDetector}.

\let\thefootnote\relax\footnotetext{\small $^*$Corresponding author.
\\ 
\texttt{\small Preprint submitted to Elsevier.} 
\hfill \textit{October 13, 2026} \\
}
\end{abstract}


\section{Motivation and significance}

The use of LLMs has become increasingly prevalent in various aspects of modern life. However, this widespread adoption also presents challenges in distinguishing between AI-generated and human-written text. This challenge is particularly important today, as verifying information and ensuring content authenticity have become increasingly urgent. For instance, in higher education, students may misuse AI tools to complete assignments and essays, potentially undermining their motivation to learn and their critical thinking skills. More concerningly, malicious actors may exploit AI tools on social media platforms to generate fake news or harmful content, thereby influencing public perception and behavior \cite{gehrmann-etal-2019-gltr}. Therefore, developing a tool for detecting AI-generated text is essential to help ensure information authenticity and prevent the spread of unreliable content.

Considerable efforts have been devoted to detecting AI-generated text, leading to the development of numerous studies, methods, and both open-source and commercial tools, such as Binoculars~\cite{hans2024binoculars}, RadarTester~\cite{hu2023radar}, DetectGPT~\cite{mitchell2023detectgpt},  Ghostbuster~\cite{verma2024ghostbuster}, GPTZero\footnote{\label{fn:gptzero}https://gptzero.me}, Turnitin\footnote{\label{fn:turnitin}https://www.turnitin.com}, and CNKI-AIGC\footnote{\label{fn:cnki}https://cx.cnki.net/}. However, most of these efforts have focused on widely spoken languages such as English, Spanish, Chinese, and Japanese. In contrast, research on detecting AI-generated text in other languages, particularly Vietnamese, remains limited~\cite{nguyen2025vietbinoculars}. This highlights the need to develop dedicated tools for detecting Vietnamese AI-generated text to address the growing demand from Vietnamese users and help ensure content authenticity.
 
Motivated by these challenges, we developed VietAIDetector, a tool specifically designed for detecting Vietnamese AI-generated text. The tool features a user-friendly interface and scientific reporting capabilities to assist users in distinguishing between AI-generated and human-written text, thereby helping ensure information authenticity. The primary goal of the tool is to support higher education by offering educators an easily deployable system to screen student submissions for potential AI misuse, providing consultative supporting evidence rather than definitive proof to uphold academic integrity.

Unlike traditional methods, VietAIDetector employs a zero-shot approach based on the Binoculars and VietBinoculars research, eliminating the need for fine-tuning or retraining language models. This approach enables VietAIDetector to detect AI-generated text without requiring domain-specific training data, thereby reducing development time and cost. This advantage is particularly important given the rapid development of new LLMs and the frequent updates to existing ones, which make fine-tuning or retraining increasingly impractical. Furthermore, VietAIDetector is built upon a pair of Vietnamese-specific language models, PhoGPT-4B and PhoGPT-4B-Chat~\cite{nguyen2024phogpt}, improving its performance in detecting Vietnamese AI-generated text compared to existing methods primarily designed for English~\cite{nguyen2025vietbinoculars}.
Table~\ref{tab:performance} summarizes the feature comparison between VietAIDetector and existing tools and methods, showing that VietAIDetector provides the most comprehensive set of operational features among the systems.

\begin{table}[!htb]
\centering
\caption{Feature comparison between VietAIDetector and existing tools and methods.}
\resizebox{\textwidth}{!}{
\label{tab:performance}
\begin{tabular}{lccccccc}
\hline
\thead{\textbf{Method/}\\\textbf{Tool}} &
\thead{\textbf{Open}\\\textbf{Source}} &
\thead{\textbf{Zero-}\\\textbf{Shot}} &
\thead{\textbf{GUI}} &
\thead{\textbf{Optimized for}\\\textbf{Vietnamese}} &
\thead{\textbf{Handles}\\\textbf{Long Texts}} &
\thead{\textbf{Adversarial}\\\textbf{Robustness}} &
\thead{\textbf{Input}\\\textbf{Formats}}\\
\hline
VietAIDetector & \ding{51} & \ding{51} & \ding{51} & \ding{51} & \ding{51} & \ding{51}\textsuperscript{a} & \ding{51} \\
Binoculars~\cite{hans2024binoculars} & \ding{51} & \ding{51} & \ding{51} & \ding{55} & \ding{55} & \ding{51} & \ding{55} \\
GLTR~\cite{gehrmann-etal-2019-gltr} & \ding{51} & \ding{51} & \ding{51} & \ding{55} & \ding{55} & \ding{55} & \ding{55} 
\\
Rank/LogRank~\cite{gehrmann-etal-2019-gltr,solaiman2019releasestrategiessocialimpacts} & \ding{51} & \ding{51} & \ding{55} & \ding{55} & \ding{55} & \ding{55} & \ding{55}
\\
RADAR~\cite{hu2023radar} & \ding{51} & \ding{55} & \ding{51} & \ding{55} & \ding{55} & \ding{55} & \ding{55} \\
DetectGPT~\cite{mitchell2023detectgpt} & \ding{51} & \ding{51} & \ding{55} & \ding{55} & \ding{55} & \ding{55} & \ding{55} \\
Ghostbuster~\cite{verma2024ghostbuster} & \ding{51} & \ding{55} & \ding{51} & \ding{55} & \ding{55} & \ding{55} & \ding{55} \\
Likelihood~\cite{solaiman2019releasestrategiessocialimpacts} & \ding{51} & \ding{51} & \ding{55} & \ding{55} & \ding{55} & \ding{55} & \ding{55}\\
OpenAI-RoBERTa~\cite{solaiman2019releasestrategiessocialimpacts} & \ding{51} & \ding{55} & \ding{55} & \ding{55} & \ding{55} & \ding{55} & \ding{55}
\\
GPTZero\footref{fn:gptzero}~\cite{adam2026gptzeroro} & \ding{55} & \ding{55} & \ding{51} & \ding{51} & \ding{51} & \ding{51} & \ding{51} \\
Turnitin\footref{fn:turnitin} & \ding{55} & \ding{55} & \ding{51} & \ding{55} & \ding{51} & \ding{51} & \ding{51} \\
CNKI-AIGC\footref{fn:cnki} & \ding{55} & \ding{55} & \ding{51} & \ding{55} & \ding{51} & \ding{51} & \ding{51} \\
\hline
\end{tabular}
}
\par\smallskip
\begin{minipage}{\textwidth}
\footnotesize\textsuperscript{a} For VietAIDetector, this marker refers to evaluation on selected AI-generated rewriting conditions in Section~\ref{sec:rewriting_robustness}. It does not indicate robustness to human-edited text or adaptive attacks.
\end{minipage}
\end{table}

The main contributions of this work are summarized as follows:
\begin{itemize}
\item Introduction of VietAIDetector, an open-source tool providing an intuitive Gradio interface for Vietnamese AI text detection across raw text, DOCX files, and PDF documents.
\item Software integration of the zero-shot VietBinoculars framework, eliminating the need for domain-specific retraining and labeled data collection.
\item A sliding-window chunking and aggregation pipeline that addresses large language model context limits and mitigates threshold drift on long documents.
\item Configurable operating thresholds across Youden's J statistic, Closest Point, and low false-positive criteria, easily maintained in \texttt{settings.py}.
\item Extensive empirical evaluations characterizing chunking stability, OCR degradation, rewriting resilience, and computational runtime throughput.
\end{itemize}

\section{Software description}

\subsection{VietAIDetector method}

The detection technique employed in VietAIDetector adopts the zero-shot scoring metric established in the VietBinoculars framework~\cite{nguyen2025vietbinoculars}, utilizing PhoGPT-4B as the observer model and PhoGPT-4B-Chat as the performer model. It first computes \texttt{log(perplexity)}, which measures the surprise of the input text relative to the performer model. Next, it calculates \texttt{cross-perplexity} to measure the divergence between the observer's expectations and the performer's predictions. Finally, a detection score is computed as the ratio of \texttt{log(perplexity)} to \texttt{cross-perplexity}, and the input text is classified as AI-generated or human-written using thresholds derived from Youden's J statistic~\cite{youden1950index}, the Closest Point~\cite{PerkinsNeilJ.2006TIo}, or TPR at 0.05 FPR on Vietnamese datasets. The updated thresholds are provided in~\ref{appendix:update-thresholds}. The algorithm is presented in algorithm~\ref{alg:vietbinoculars}.

\begin{algorithm}[!ht]
\small
\DontPrintSemicolon
\caption{VietBinoculars score computation and AI/Human decision rule}
\label{alg:vietbinoculars}

\SetKwInOut{Input}{Input}
\SetKwInOut{Output}{Output}
\SetKwProg{Fn}{Function}{:}{}
\SetKwFunction{ComputeScore}{ComputeVietBinocularsScore}
\SetKwFunction{Classify}{ClassifyText}

\textbf{Part 1: Compute the VietBinoculars score}\;
\Fn{\ComputeScore{$s$, $M_1$, $M_2$}}{
    \Input{Raw string $s$; observer model $M_1$ (PhoGPT-4B); performer model $M_2$ (PhoGPT-4B-Chat); shared BPE tokenizer\cite{nguyen2020phobert,trieu2022clustering}}
    \Output{VietBinoculars score $B_{M_1,M_2}(s)$}
    $\vec{x} \leftarrow \text{BPE}(s)$\;
    $L \leftarrow$ number of tokens in $\vec{x}$\;
    $Y \leftarrow M_1(\vec{x})$ \tcp*{next-token prediction distributions of the observer model}
    $Z \leftarrow M_2(\vec{x})$ \tcp*{next-token prediction distributions of the performer model}
    $\log \mathrm{PPL}_{M_2}(s) \leftarrow -\dfrac{1}{L}\displaystyle\sum_{i=1}^{L} \log\left(Z_{i,x_i}\right)$\;
    $\log \text{X-PPL}_{M_1,M_2}(s) \leftarrow -\dfrac{1}{L}\displaystyle\sum_{i=1}^{L} Y_i \cdot \log\left(Z_i\right)$\;
    $B_{M_1,M_2}(s) \leftarrow \dfrac{\log \mathrm{PPL}_{M_2}(s)}{\log \text{X-PPL}_{M_1,M_2}(s)}$\;
    \KwRet{$B_{M_1,M_2}(s)$}
}
\BlankLine
\textbf{Part 2: Classify the text as human-written or AI-generated}\;
\Fn{\Classify{$B_{M_1,M_2}(s)$, $t^*$}}{
    \Input{VietBinoculars score $B_{M_1,M_2}(s)$; threshold $t^*$, chosen from Youden's J statistic, the Closest Point approach, or TPR@0.05FPR, and derived from the Vietnamese training datasets \cite{nguyen2025vietbinoculars} and updated in~\ref{appendix:update-thresholds}} 
    \Output{Predicted label $\in \{\text{Human}, \text{AI}\}$}
    \uIf{$B_{M_1,M_2}(s) \geq t^*$}{
        $\mathrm{label} \leftarrow \text{Human}$ \tcp*{human-written text tends to yield a higher score}
    }
    \uElse{
        $\mathrm{label} \leftarrow \text{AI}$ \tcp*{AI-generated text tends to yield a lower score}
    }
    \KwRet{$\mathrm{label}$}
}
\end{algorithm}

Using the aforementioned approach, VietAIDetector achieves state-of-the-art performance on out-of-domain Vietnamese datasets, as demonstrated in previous research~\cite{nguyen2025vietbinoculars}. The option to use a threshold based on TPR at 0.05 FPR minimizes false alarms. This is crucial in higher education, where misclassifying AI-generated text may have serious ethical and legal consequences. Furthermore, VietAIDetector is designed to handle long input texts and various prompting strategies that may attempt to bypass detection systems. 

\subsection{Software architecture}
\begin{figure}[!htb]
    \centering
    \includegraphics[width=1.1\textwidth]{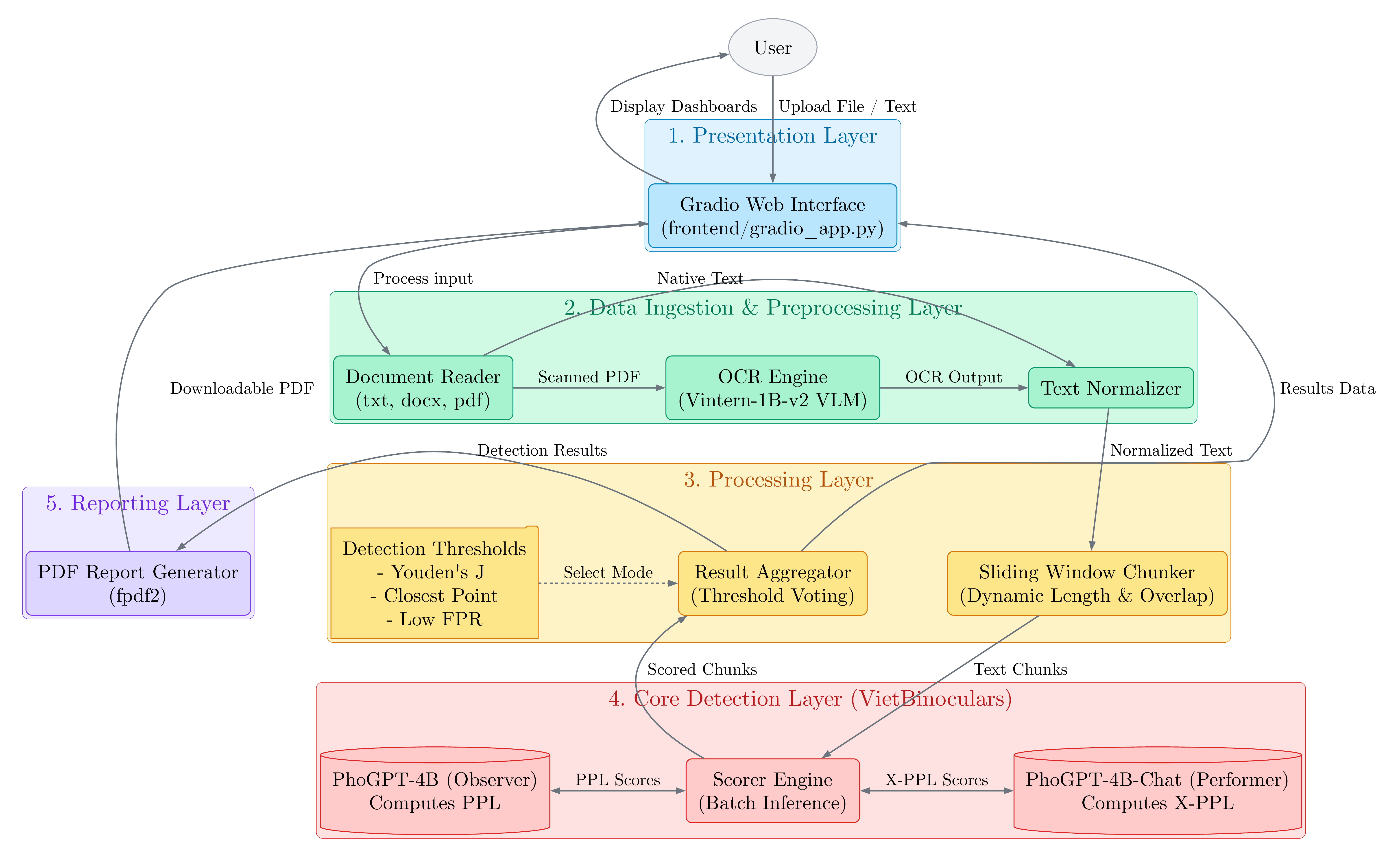}
    \caption{The overall architecture of the VietAIDetector tool.}
    \label{fig:architecture}
\end{figure}

The overall architecture of VietAIDetector is illustrated in Figure~\ref{fig:architecture}. It consists of five main layers. The \texttt{Presentation Layer} provides a Gradio-based web interface that accepts raw text or file uploads and allows users to configure detection parameters such as the threshold mode and chunk size. The \texttt{Data Ingestion and Preprocessing Layer} parses input documents (\texttt{.txt}, \texttt{.docx}, or native PDF), routes scanned PDFs to an OCR engine based on the Vintern-1B-v2 vision-language model~\cite{doan2024vintern1b}, and normalizes the extracted text for compatibility with the downstream language models. The \texttt{Processing Layer} uses a sliding-window chunker to divide long documents into overlapping chunks while an aggregator combines the chunk-level outputs into document-level statistics. The \texttt{Core Detection Layer} implements the VietBinoculars algorithm by using PhoGPT-4B as the observer model and PhoGPT-4B-Chat as the performer model to compute the detection score for each chunk and classify it as AI-generated or human-written according to the selected threshold. Finally, the \texttt{Reporting Layer} aggregates the chunk-level classification results into an overall decision and generates a detailed, downloadable PDF report with color-coded highlights for each chunk.
The five main layers correspond to the \texttt{frontend}, \texttt{preprocessing}, \texttt{processing}, \texttt{core}, and \texttt{reporting} modules in the VietAIDetector source code. In addition, the \texttt{config} module centralizes shared configuration parameters, including model names, detection thresholds, chunking settings, and device allocation. The \texttt{schemas} module defines the structured data containers for exchanging preprocessing outputs, chunk-level scores, and document-level detection results across the processing pipeline. Together, these modules form a modular architecture that facilitates maintenance, future upgrades, and system extensibility.

\subsection{Software functionalities}
\subsubsection{Multi-format input and text preprocessing}
In the \texttt{preprocessing} and \texttt{frontend} modules, VietAIDetector supports both direct text input and file upload workflows. The document reader accepts \texttt{.txt}, \texttt{.docx}, and \texttt{.pdf} files and routes each format to the appropriate extraction pipeline. Plain text and DOCX documents are parsed using built-in text readers and the \texttt{python-docx} library, respectively, while PDF files are first checked for a native text layer. If no text layer is detected, an OCR engine based on the \texttt{Vintern-1B-v2} vision-language model is used to extract text from the scanned document. Otherwise, text is extracted directly from native PDF files using the \texttt{PyMuPDF}\footnote{https://pymupdf.readthedocs.io} library. 

Vietnamese scanned documents often present challenges for OCR because of linguistic characteristics such as diacritics and ligatures. Therefore, the multimodal vision-language model Vintern-1B-v2 is employed for document transcription. Because transcription noise and potential hallucinated text in vision-language models can significantly perturb downstream detection scores, the OCR stage adopts deterministic decoding settings together with a structured extraction prompt (see \texttt{OCR\_PROMPT} in \texttt{settings.py}). To reduce startup time and improve VRAM efficiency, the OCR engine is loaded on demand only when a scanned PDF is detected and is executed on a single GPU.

Moreover, the extracted text is normalized through de-hyphenation, line-break restoration, and whitespace normalization. The preprocessing stage also removes noisy and excessively short paragraphs before the detection process.

\subsubsection{Sliding-window chunking for long documents}

LLMs have a limited context size, and exceeding this limit may degrade detection performance because of statistical feature dilution or attention degradation in excessively long sequences. Rather than truncating the input, long documents are divided into overlapping token chunks using configurable window and overlap sizes. This design preserves contextual continuity while ensuring that every chunk remains within the token limit of the employed LLMs.

The input text sequence $S = (x_1, x_2, \dots, x_N)$ with $N$ tokens (encoded using a BPE tokenizer) is divided into $K$ overlapping chunks using a sliding window of size $W$ (where $W \le L_{max}$) and stride $D$. Here, $L_{max}$ denotes the maximum effective context length supported by the employed LLMs, and it is assumed that $N \gg L_{max}$. The $k$-th chunk, denoted by $C_k$, is given by
\[
    C_k = (x_{1 + (k-1)D}, \dots, x_{W + (k-1)D}).
\]

The total number of chunks $K$ is calculated as
\[
    K = \left\lfloor \frac{N - W}{D} \right\rfloor + 1.
\]
To handle short trailing segments, let $m$ denote the minimum admissible chunk length and let $\ell_K = |C_K|$. The following post-processing rule is then applied:
\[
(\widetilde{C},\widetilde{K})=
\begin{cases}
\left(\{C_1,\dots,C_{K-2},C'_{K-1}\},\,K-1\right), & K>1,\ \ell_K<m,\\
\left(\{C_1,\dots,C_K\},\,K\right), & \text{otherwise.}
\end{cases}
\]
\[
C'_{K-1} = (x_{1+(K-2)D},\dots,x_N).
\]
Consequently, all downstream computations are performed on the $\widetilde{K}$ resulting chunks, reducing high-variance estimates caused by very short trailing segments while preserving complete document coverage. For each retained chunk in $\widetilde{C}$, the local VietBinoculars score is obtained using
\[
B_{M_1, M_2}(C_k) = \frac{\log \mathrm{PPL}_{M_2}(C_k)}{\log \text{X-PPL}_{M_1, M_2}(C_k)}.
\]
The chunk-level decision then follows algorithm~\ref{alg:vietbinoculars}: \texttt{AI} if $B_{M_1,M_2}(C_k)<t^*$; otherwise, \texttt{Human}. The complete implementation of the chunking process is provided in \texttt{chunker.py} within the \texttt{processing} module.

\subsubsection{Detection score and configurable decision thresholds}

For each chunk, the \texttt{core} module computes the VietBinoculars score from the ratio of perplexity to cross-perplexity using PhoGPT-4B and PhoGPT-4B-Chat. In \texttt{scorer.py}, both models are loaded in evaluation mode and distributed across two GPUs, when available, to balance VRAM usage and improve throughput. A shared tokenizer generates a single batched encoding, which is then forwarded through both models to obtain the observer and performer logits.

To ensure memory-safe inference, gradient-free execution, \texttt{bfloat16} precision, capped input length, and fixed-size chunk batching are employed. CUDA streams are also synchronized before cross-device score composition. Together, these implementation choices provide stable dual-GPU scoring for long documents without causing out-of-memory errors.

Moreover, the software provides multiple operating thresholds, including Youden's J statistic, the Closest Point approach, and a low-FPR criterion. Users can select the option that best matches their preferred trade-off between sensitivity and false alarms. These options are configured in \texttt{settings.py} and can be selected through the Gradio interface. The final chunk-level decision is obtained by comparing the detection score with the selected threshold.

\subsubsection{Chunk-level labeling and document-level aggregation}
\label{sec:document_aggregation}

Each retained chunk is classified as \texttt{AI} or \texttt{Human}, and the document-level decision is obtained through majority voting~\cite{kuncheva04combining}. This strategy assumes that individual chunks can provide complementary evidence about the origin of the document and is expressed as
\[
\text{Vote}(S) = \sum_{k=1}^{\widetilde{K}}\mathbb{I}(B_{M_1, M_2}(C_k) < t^*),
\]
where $\mathbb{I}\{\cdot\}$ is the indicator function. The percentage of chunks classified as AI-generated is then computed as
\[ P_\text{AI} = \frac{\text{Vote}(S)}{\widetilde{K}} \times 100\%. \]
Finally, the document is assigned one of the following labels based on $P_\text{AI}$:
\[\text{Decision}(S) = \begin{cases} 
\textit{AI-generated} & \text{if } P_\text{AI} > 50\% \\    
\textit{Human-written but contains AI parts} & \text{if } 0\% < P_\text{AI} \leq 50\% \\
\textit{Human-written} & \text{if } P_\text{AI} = 0\% 
\end{cases}\]
Document-level aggregation based on the majority voting strategy is implemented in \texttt{aggregator.py} within the \texttt{processing} module.

\subsubsection{User interface, reporting and system integration}
VietAIDetector integrates user interaction, result reporting, and reliability control into a unified workflow. Through the Gradio-based \texttt{frontend} module, users can upload documents, select decision thresholds, and configure chunking parameters (window size and overlap) without modifying the source code. Progress bars and status messages provide real-time feedback throughout the detection process.

After inference, the system automatically generates a downloadable PDF report containing summary statistics and color-coded chunk-level outcomes, supporting transparent inspection, record keeping, and post hoc analysis. The report is generated using the \texttt{fpdf2}\footnote{https://py-pdf.github.io/fpdf2/} library. The implementation is provided in \texttt{pdf\_report.py} within the \texttt{reporting} module. 

To improve decision reliability during practical use, the pipeline also incorporates minimum-token constraints, extraction-status verification, and user-facing warning messages. These safeguards reduce the risk of producing predictions from unsuitable or malformed inputs.

The software also provides JSON output for integration with external systems and automated workflows. Implemented through the \texttt{schemas} module, it serializes the \texttt{Detection\allowbreak Result} and \texttt{ChunkDetail} data models into a hierarchical JSON structure containing document-level metadata and chunk-level detection results. 

\section{Illustrative examples}
\subsection{Examples using text input}

\begin{figure}[!htb]
    \centering
    \includegraphics[width=1.0\textwidth]{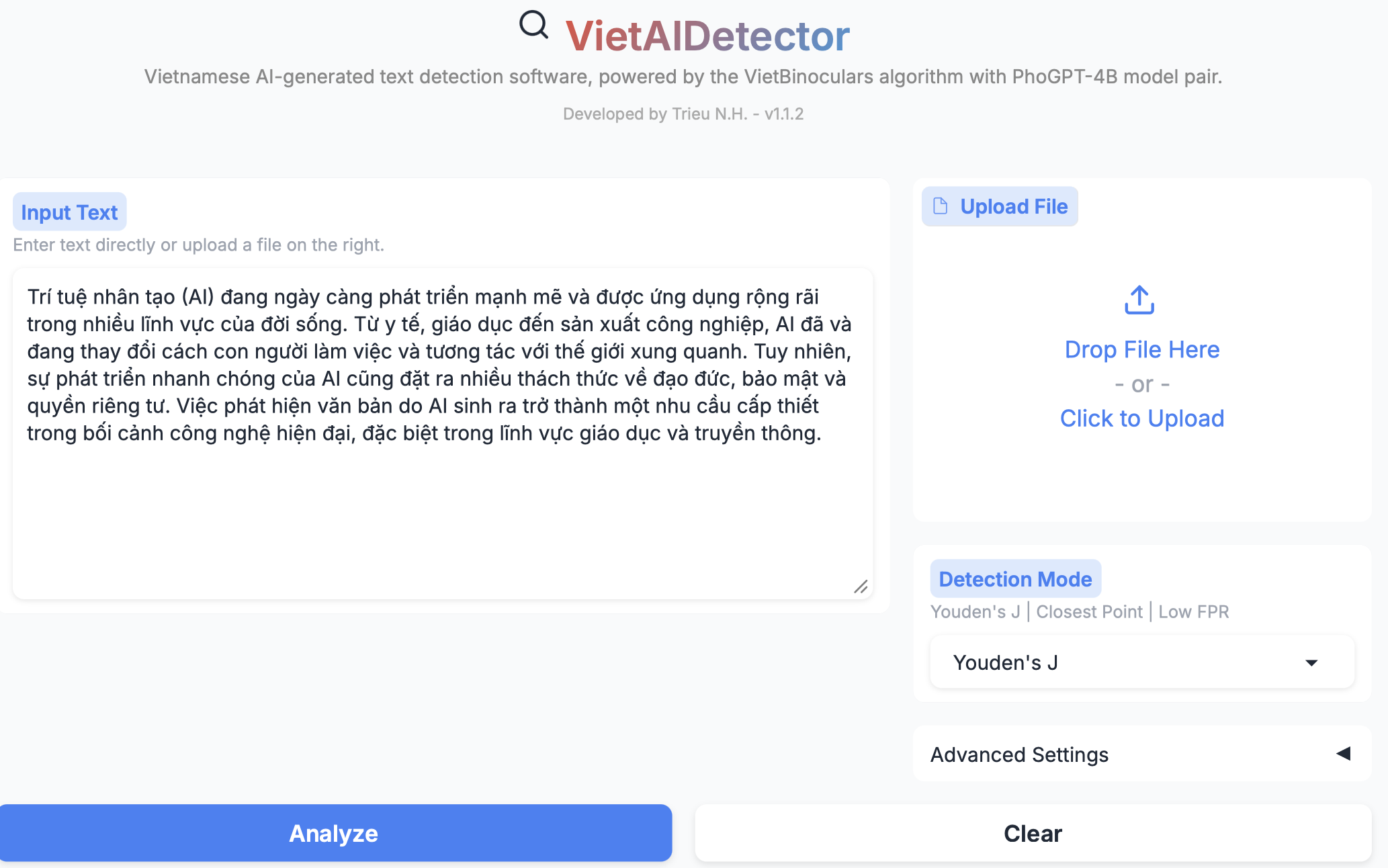}
    \caption{The VietAIDetector interface for AI-generated text detection using raw text input.}
    \label{fig:example_raw_1_main}
\end{figure}

This section demonstrates the AI-generated text detection functionality of VietAIDetector using raw text input. Users can enter text into the \texttt{Input Text} field and click \texttt{Analyze} to start the detection process, as illustrated in Figure~\ref{fig:example_raw_1_main}. 
Example inputs can also be obtained from out-of-domain datasets, including News articles\footnote{https://doi.org/10.57967/hf/6233} and Literary works\footnote{https://doi.org/10.57967/hf/6234}. Table~\ref{tab:example_raw_1_main} summarizes the corresponding fields for human-written and AI-generated text. 
\begin{table}[!htb]
\centering
\caption{Field names used to extract human-written and AI-generated text from the evaluation datasets.}
\label{tab:example_raw_1_main}
\resizebox{\textwidth}{!}{
\begin{tabular}{llcr}
\toprule
\multirow{2}{*}{\textbf{Domain}} & \multirow{2}{*}{\textbf{Dataset file}} & \multicolumn{2}{c}{\textbf{Field names}}\\
  \cmidrule(lr){3-4} 
  & & \multicolumn{1}{c}{\textbf{Human}} & \multicolumn{1}{c}{\textbf{AI}} \\
\midrule
News articles & test & \texttt{text} & \texttt{google-gemma-3-12b-it-generated\_text} \\
\midrule
\multirow{2}{*}{Literary works} & Gemma-3-12B-VuTrongPhung & \texttt{text} & \texttt{google-gemma-3-12b-it-hf\_generated\_text\_wo\_prompt}\\
& Sailor2-8B-VuTrongPhung & \texttt{text} & \texttt{sail-Sailor2-8B-Chat-hf\_generated\_text\_wo\_prompt} \\ 
\bottomrule
\end{tabular}
}
\end{table}

\begin{figure}[!htb]
    \centering
    \includegraphics[width=1.0\textwidth]{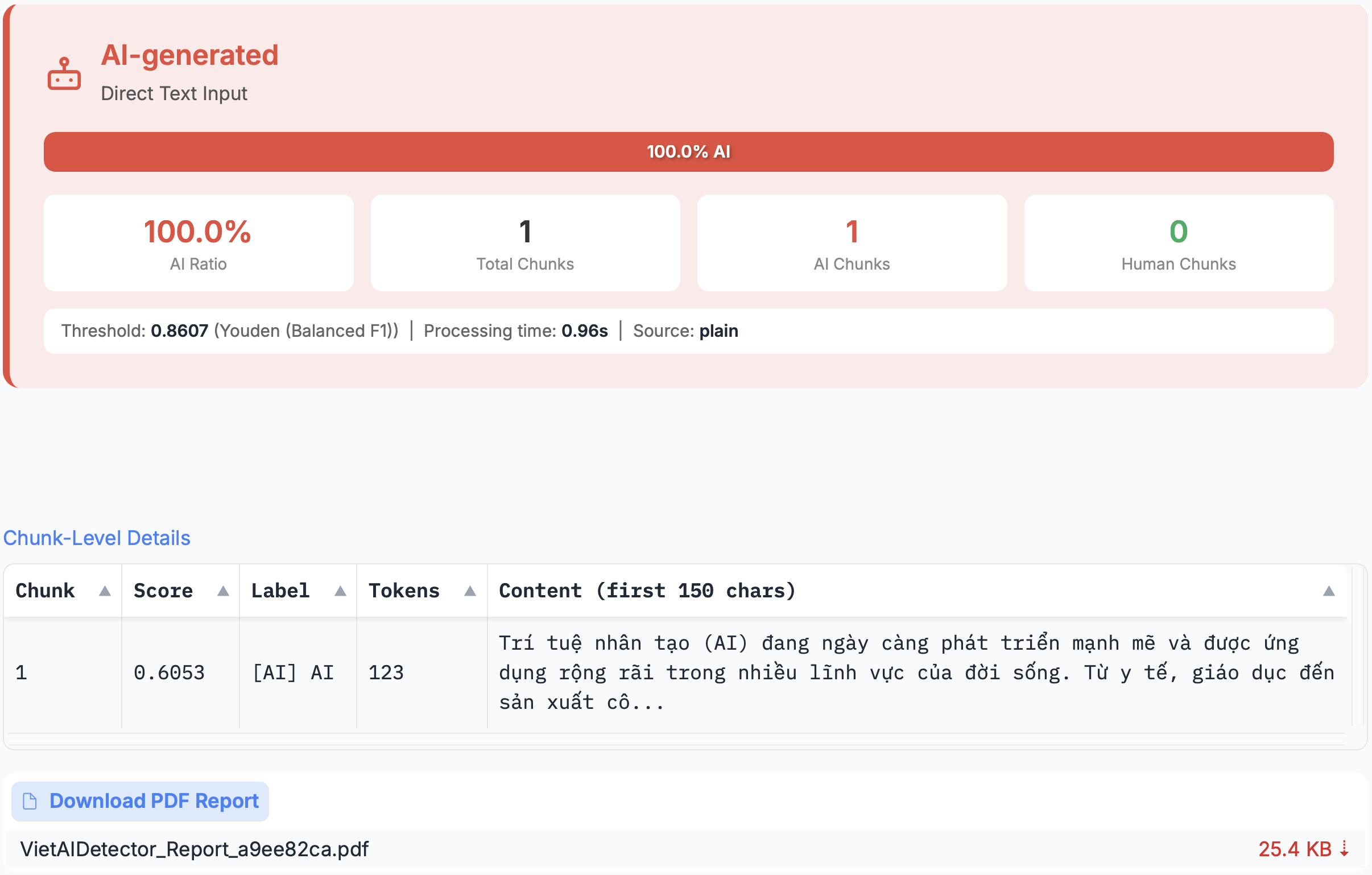}
    \caption{The VietAIDetector interface showing the detection results for the input text.}
    \label{fig:example_raw_1_result}
\end{figure}

Figure~\ref{fig:example_raw_1_result} shows the interface after the detection process is completed. In addition to the document-level prediction, the interface presents chunk-level details, including the chunk index, detection score, predicted label, token count, and chunk content. The \texttt{Download} button (\textcolor{red}{$\downarrow$}) allows users to export the complete detection report as a PDF.

\subsection{Examples using file upload and parameter configuration}

\begin{figure}[!htb]
\centering
\includegraphics[width=1.0\textwidth]{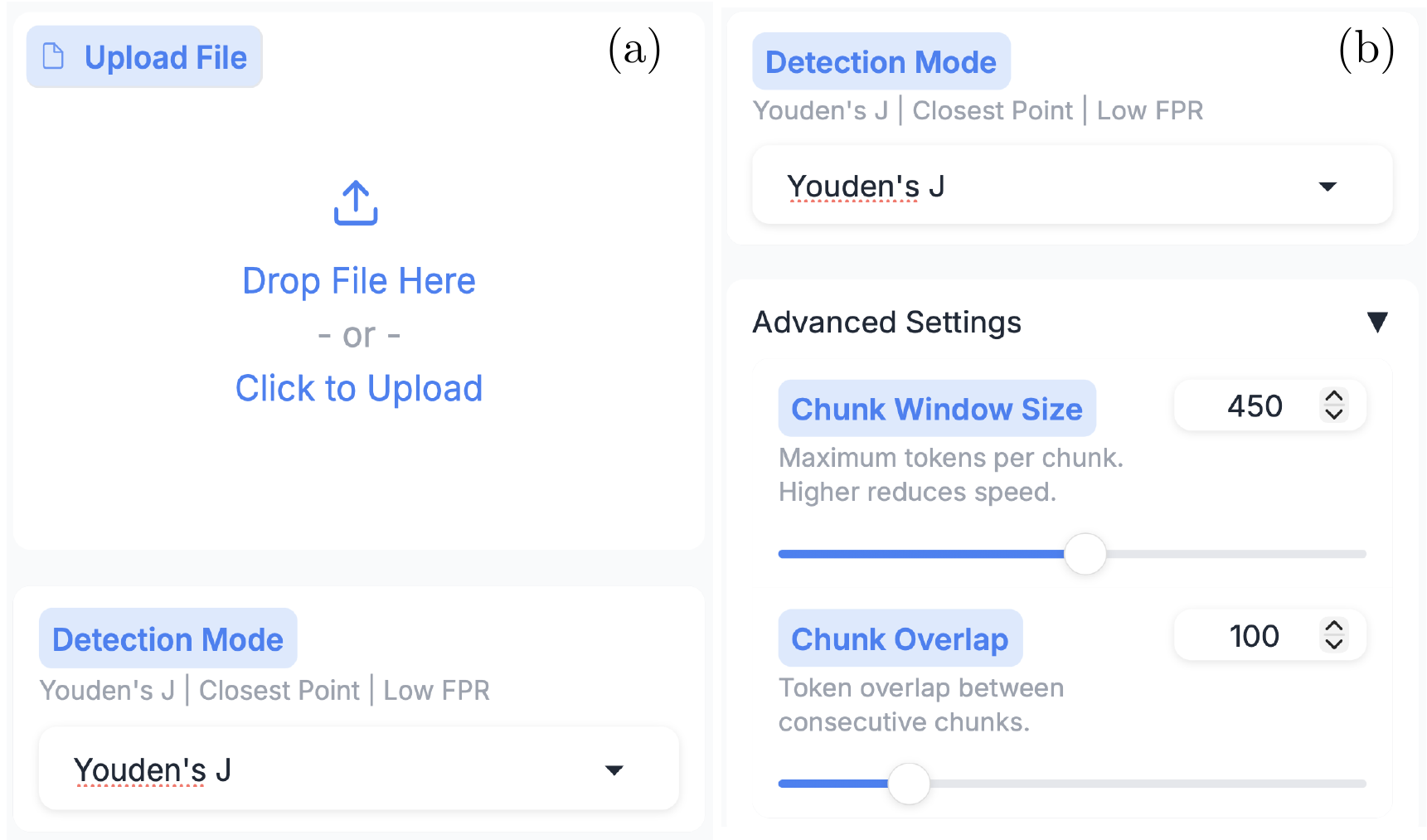}
\caption{The VietAIDetector interface for file upload and detection parameter configuration.}
\label{fig:example_upload_configuration}
\end{figure}

The file upload and parameter configuration process is illustrated in this section. Example files, including text documents, native PDFs, and scanned PDFs, are provided in the \texttt{examples} folder of the source code. Files can be uploaded either by drag-and-drop or by clicking the upload area. Figure~\ref{fig:example_upload_configuration}(a) shows the file upload interface, while Figure~\ref{fig:example_upload_configuration}(b) presents the parameter configuration interface, including the detection threshold and chunking parameters used during the detection process. Note that these configuration parameters are also included in the downloaded PDF report, enabling other users to reproduce and verify the detection results.

\subsection{Exploratory benchmark using the programmatic API}
 
This exploratory benchmark examines the effects of window size $W$ and overlap $O$ and illustrates long-document processing using news datasets available as of July 2026. We name these datasets Claude-Sonnet-4.6-News, Gemini-3.7-Flash-News, and GPT-5.6-Luna-News. Each contains 20 long AI-generated documents from the model identified in its name. The datasets cover economics, politics, culture, society and sports. GPTZero was used as a baseline for comparison due to its functional similarity to VietAIDetector (see Table~\ref{tab:performance}). Evaluations were performed in August 2026 through the official GPTZero Web Dashboard using Model 4.9b (August 12, 2026 release). Unabridged texts preserved under \texttt{benchmarks\allowbreak\allowbreak/datasets} were submitted directly without manual modification or length truncation, and the resulting prediction records are archived in \texttt{benchmarks\allowbreak/results\allowbreak/GPTZero}. The benchmark can be reproduced using \texttt{benchmarks\allowbreak/run\_eval.py} in the source code.

\begin{figure}[!htb]
    \centering
    \includegraphics[width=1.0\textwidth]{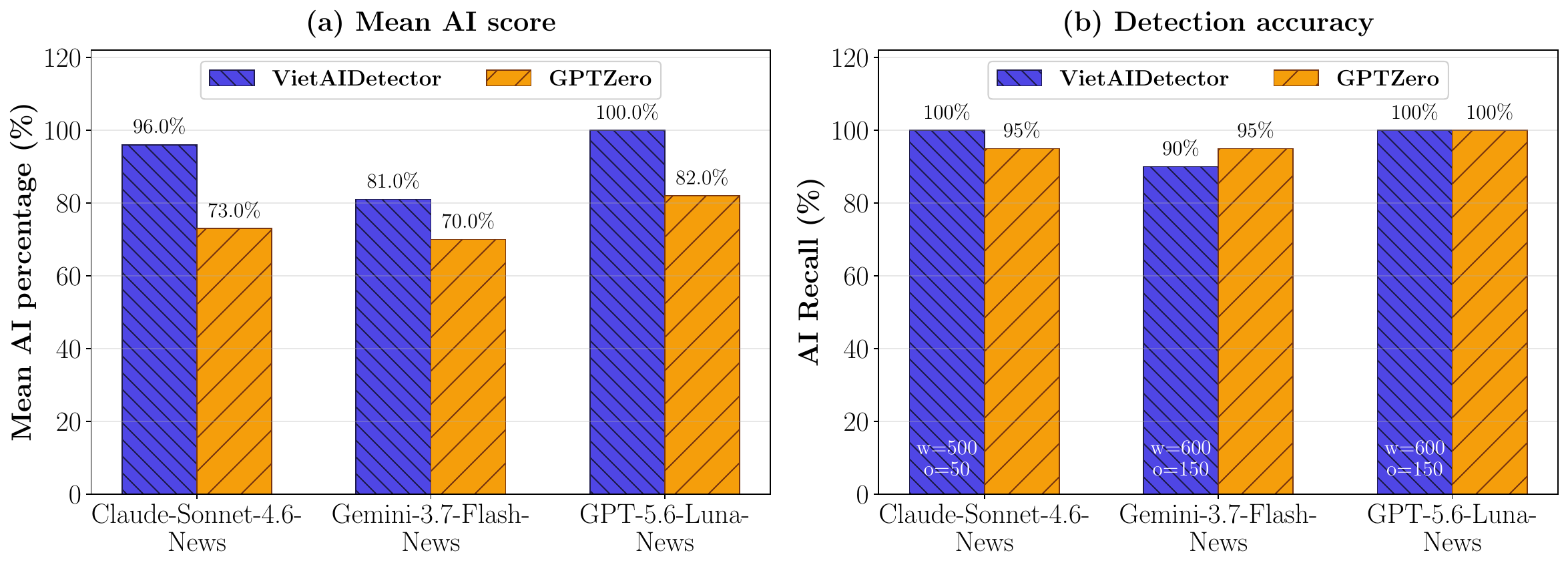}
    \caption{Exploratory comparison on Claude-Sonnet-4.6-News, Gemini-3.7-Flash-News, and GPT-5.6-Luna-News, with 20 documents per dataset. VietAIDetector configurations were selected using these same datasets, so the results do not provide an independent estimate of generalization performance. (a) Average AI score. (b) AI recall under majority voting with configuration $(W=500, O=50)$ and $(W=600, O=150)$.}
    \label{fig:benchmark_api_usage}
\end{figure}

Figure~\ref{fig:benchmark_api_usage} summarizes the observed results for configurations selected after this survey. These exploratory results do not constitute an independent evaluation of generalization performance. Figure~\ref{fig:benchmark_api_usage}(a) illustrates that, on these exploratory samples, VietAIDetector yielded a higher average AI score than GPTZero. Notably, on Gemini-3.7-Flash-News, VietAIDetector achieved an average AI score of 0.81, compared with 0.7 for GPTZero, while GPTZero achieved higher AI recall. This discrepancy arises because GPTZero considers additional conditions beyond whether the AI probability is greater than the $50\%$ threshold, whereas VietAIDetector bases its decisions solely on the AI score threshold.

The configuration survey evaluated window sizes from 200 to 650 and overlap sizes from 50 to 150, both in steps of 50. Table~\ref{tab:grid_summary_compact} in~\ref{appendix:grid_search} reports all configurations. Because configuration selection and evaluation used the same small datasets, the selected results may be optimistic.

\subsection{Selection and evaluation of chunking parameters on news datasets}
\label{sec:chunking_evaluation}

To evaluate chunking independently of configuration selection, we constructed Gemma-3-12B-News-Concat-Dev and Gemma-3-12B-News-Concat-Test from Gemma-3-12B-News~\cite{nguyen2025vietbinoculars}. The source corpus contains 7,222 paired Human news passages and Gemma-3-12B-it continuations generated through greedy decoding. The two sets share the news domain but serve separate development and test roles.

Before splitting, we grouped candidate passage pairs sharing source article IDs, prompts, or text content with five-word Jaccard similarity of at least 0.8. We excluded groups linked to threshold-calibration data and retained one eligible pair per remaining group. The resulting 6,899 passage pairs were randomly divided into development and test partitions in a 20/80 ratio before concatenation. Within each partition, we joined three to seven passages separately for Human and AI, using the same sources and order for each pair. Both documents in each pair contained at least 2,048 BPE tokens after preprocessing. The final sets contain 296 development pairs and 1,179 test pairs, without source reuse across pairs or partitions. This protocol ensures that configuration selection remains strictly decoupled from test evaluation, preventing direct source reuse and substantially mitigating cross-partition leakage. Table~\ref{tab:concat_news_datasets} summarizes their size and token-length distributions. \ref{appendix:token_length_distributions} and Figure~\ref{fig:token_length_distribution} present the empirical token-length probability distributions across all datasets.

\begin{table}[!ht]
\centering
\caption{Summary and token-length distributions of Gemma-3-12B-News-Concat-Dev and Gemma-3-12B-News-Concat-Test. The number of documents (\#Docs) is evenly divided between Human and AI classes. PhoGPT BPE token counts are measured after preprocessing, without truncation or special tokens. SD denotes sample standard deviation, and brackets report the first and third quartiles [Q1, Q3].}
\label{tab:concat_news_datasets}
\resizebox{\textwidth}{!}{%
\begin{tabular}{llrcccc}
\toprule
\multirow{2}{*}{\textbf{Dataset}} & \multirow{2}{*}{\textbf{Role}} & \multirow{2}{*}{\textbf{\#Docs}} & \multicolumn{2}{c}{\textbf{Human tokens/doc}} & \multicolumn{2}{c}{\textbf{AI tokens/doc}} \\
\cmidrule(lr){4-5}\cmidrule(lr){6-7}
& & & \textbf{Mean $\pm$ SD} & \textbf{Median [Q1, Q3]} & \textbf{Mean $\pm$ SD} & \textbf{Median [Q1, Q3]} \\
\midrule
Gemma-3-12B-News-Concat-Dev & Development & 592 & 2,386.4 $\pm$ 240.2 & 2,369.0 [2,191.8, 2,507.2] & 2,712.4 $\pm$ 411.1 & 2,666.0 [2,396.8, 2,996.2] \\
Gemma-3-12B-News-Concat-Test & Test & 2,358 & 2,401.4 $\pm$ 255.2 & 2,362.0 [2,206.5, 2,524.0] & 2,703.3 $\pm$ 422.3 & 2,646.0 [2,382.0, 2,957.5] \\
\bottomrule
\end{tabular}}
\end{table}

We evaluated 40 configurations with $W\in\{200,250,\ldots,650\}$ and $O\in\{0,50,100,150\}$ on the development set. Configuration selection maximized macro-F1 under majority voting and the updated Youden threshold, with ties resolved by AUROC, fewer mean chunks, and then smaller $W$ and $O$. Among configurations tied at macro-F1 and AUROC of 1.0000, $(650,150)$ and $(600,0)$ had the fewest mean chunks within the overlapping and non-overlapping groups, respectively. Figure~\ref{fig:development_chunking_selection} in~\ref{appendix:development_selection} presents all development configurations and identifies the selected candidates. We fixed these configurations before test evaluation and compared them with full-document scoring and truncation at 768 tokens. The calibration thresholds in~\ref{appendix:update-thresholds} remained unchanged. The main analysis uses the Youden threshold of approximately 0.928.

Evaluation was performed at the document level, with AI as the positive class. Majority voting predicts AI when more than half of the chunks have scores below the fixed threshold $t^*$, whereas mean aggregation compares the average chunk score with that threshold. The document scores used to calculate AUROC are defined as
\begingroup
\begin{gather*}
r_{\mathrm{vote}}(S) = \frac{\text{Vote}(S)}{\widetilde{K}}, \quad
r_{\mathrm{w\text{-}vote}}(S) = \frac{\sum_{k=1}^{\widetilde{K}} w_k \mathbb{I}(B_{M_1,M_2}(C_k) < t^*)}{\sum_{k=1}^{\widetilde{K}} w_k}, \\[3pt]
r_{\mathrm{mean}}(S) = -\frac{1}{\widetilde{K}}\sum_{k=1}^{\widetilde{K}} B_{M_1,M_2}(C_k), \quad
r_{\mathrm{w\text{-}mean}}(S) = -\frac{\sum_{k=1}^{\widetilde{K}} w_k B_{M_1,M_2}(C_k)}{\sum_{k=1}^{\widetilde{K}} w_k}.
\end{gather*}
\endgroup
where $w_k$ denotes the token count of chunk $C_k$, including overlapping tokens. Higher values of these scores indicate stronger evidence of AI generation. For either single-score baseline, mean aggregation $r_{\mathrm{mean}}$ preserves the original ranking, whereas majority voting $r_{\mathrm{vote}}$ produces a binary score. Uncertainty was estimated from 2,000 bootstrap resamples of complete Human/AI source pairs. Paired randomization tests used 2,000 permutations, with Holm adjustment across 12 tests of AUROC and macro-F1.

\begin{table}[!ht]
\centering
\caption{Document-level performance on Gemma-3-12B-News-Concat-Test with 1,179 Human and 1,179 AI documents at the fixed Youden threshold ($t^* \approx 0.928$). Conditions include unsegmented full-document scoring, single-chunk truncation at 768 tokens, and sliding-window chunking $(W, O)$ with window size $W$ and overlap $O$ in tokens. Brackets show 95\% source-pair bootstrap confidence intervals. Majority AUROC uses vote fractions $r_{\mathrm{vote}}$ and $r_{\mathrm{w\text{-}vote}}$, and mean AUROC uses negative mean scores $r_{\mathrm{mean}}$ and $r_{\mathrm{w\text{-}mean}}$. For single-score baselines, majority AUROC is based on binary decisions. FP and FN denote false positives and false negatives, respectively.}
\label{tab:long_document_performance}
\resizebox{\textwidth}{!}{%
\begin{tabular}{llccrr}
\toprule
\multirow{2}{*}{\textbf{Condition}} & \multirow{2}{*}{\textbf{Aggregation}} & \textbf{AUROC} & \textbf{Macro-F1} & \multirow{2}{*}{\textbf{FP}} & \multirow{2}{*}{\textbf{FN}} \\
& & \textbf{[95\% CI]} & \textbf{[95\% CI]} & & \\
\midrule
Full document & Majority & 0.7816 [0.7668, 0.7956] & 0.7707 [0.7533, 0.7867] & 515 & 0 \\
Full document & Mean & 1.0000 [1.0000, 1.0000] & 0.7707 [0.7533, 0.7867] & 515 & 0 \\
Truncated at 768 & Majority & 0.9572 [0.9491, 0.9652] & 0.9571 [0.9490, 0.9652] & 100 & 1 \\
Truncated at 768 & Mean & 0.9998 [0.9995, 1.0000] & 0.9571 [0.9490, 0.9652] & 100 & 1 \\
$(600,0)$ & Majority & 1.0000 [1.0000, 1.0000] & 0.9992 [0.9979, 1.0000] & 2 & 0 \\
$(600,0)$ & Mean & 1.0000 [1.0000, 1.0000] & 0.9970 [0.9945, 0.9992] & 7 & 0 \\
$(650,150)$ & Majority & 1.0000 [1.0000, 1.0000] & 0.9970 [0.9949, 0.9987] & 7 & 0 \\
$(650,150)$ & Mean & 1.0000 [1.0000, 1.0000] & 0.9962 [0.9936, 0.9983] & 9 & 0 \\
$(650,150)$ & Weighted vote & 1.0000 [1.0000, 1.0000] & 0.9962 [0.9936, 0.9983] & 9 & 0 \\
$(650,150)$ & Weighted mean & 1.0000 [1.0000, 1.0000] & 0.9962 [0.9936, 0.9983] & 9 & 0 \\
\bottomrule
\end{tabular}}
\end{table}

Document-level classification results in  Table~\ref{tab:long_document_performance} show that chunking substantially improved performance at the fixed threshold. Under majority voting, $(650,150)$ and $(600,0)$ achieved macro-F1 scores of 0.9970 and 0.9992, producing only 7 and 2 false positives, respectively, with zero false negatives. Overlapping chunking exceeded truncation and full-document scoring in macro-F1 by 0.0399 and 0.2264, with bootstrap intervals and paired tests confirming statistically distinguishable gains (Holm-adjusted $p = 0.0060$). However, its difference from non-overlapping chunking was not statistically significant. Alternative aggregation functions at $(650,150)$ yielded an identical macro-F1 of 0.9962, with differences from unweighted majority voting confirming no statistical significance ($p > 0.05$), demonstrating stability across voting rules. These results validate chunking under the operational threshold while indicating that overlapping windows provide comparable rather than superior classification.

Although full-document scoring achieved an AUROC of 1.0000, it produced 515 false positives and a reduced macro-F1 of 0.7707 at the fixed threshold. This discrepancy indicates severe threshold drift on untruncated texts despite strong score separation. Consequently, chunking benefits classification primarily by preserving threshold validity across document lengths rather than improving underlying score discrimination.

Computational performance was measured on an NVIDIA A100-SXM4 GPU (40 GB) with bfloat16 precision, processing up to eight chunks per document for chunking and one document at a time for baselines. Inference ran on Python 3.12.11, PyTorch 2.8.0, Transformers 4.39.3, and CUDA 12.8. All inputs contained at most 4,466 tokens and fit within the 8,192-token context window. Thus, this evaluation compares processing strategies within supported length limits, showing that chunking substantially mitigates threshold drift on these concatenated texts even before reaching the model context boundary.

\begin{table}[!ht]
\centering
\caption{Computational performance on a single A100 GPU. Time covers document preprocessing and scoring, excluding model loading, warmup, and report generation. P95 represents the 95th percentile of processing time. Coverage is the mean proportion of source tokens retained, and peak GPU memory denotes allocated memory, including model weights.}
\label{tab:long_document_cost}
\resizebox{\textwidth}{!}{%
\begin{tabular}{lcccccc}
\toprule
\textbf{Condition} & \makecell{\textbf{Mean}\\\textbf{s/doc}} & \makecell{\textbf{P95}\\\textbf{s/doc}} & \textbf{Docs/s} & \textbf{Chunks/doc} & \makecell{\textbf{Coverage}\\\textbf{(\%)}} & \makecell{\textbf{Peak GPU}\\\textbf{memory (GiB)}} \\
\midrule
Full document & 0.813 & 1.213 & 1.23 & 1.00 & 100.00 & 17.63 \\
Truncated at 768 & 0.153 & 0.160 & 6.52 & 1.00 & 30.69 & 13.95 \\
$(600,0)$ & 0.412 & 0.511 & 2.43 & 4.64 & 100.00 & 15.03 \\
$(650,150)$ & 0.501 & 0.634 & 2.00 & 5.32 & 100.00 & 15.14 \\
\bottomrule
\end{tabular}}
\end{table}

Table~\ref{tab:long_document_cost} shows the trade-off between processing cost and input coverage. Overlapping chunking achieved a throughput of 2.00 documents per second with 15.14 GiB peak GPU memory and a 95th percentile latency of 0.634 seconds. In comparison, full-document scoring processed only 1.23 documents per second while requiring 17.63 GiB of GPU memory. Although truncation achieved the highest throughput, it retained only 30.69\% of source tokens, leaving most of the document unexamined. Non-overlapping chunking reached 2.43 documents per second with 15.03 GiB memory, maintaining full coverage with minimal computational overhead.

These findings indicate that sliding-window chunking achieves high throughput and complete input coverage on concatenated news documents while mitigating threshold drift under the evaluated conditions. However, concatenation does not reproduce the discourse coherence of naturally authored articles, topic annotations remain incomplete across source texts, and the evaluation relies on an existing news corpus~\cite{nguyen2025vietbinoculars}. Evaluating naturally authored documents, testing diverse generator architectures, and calibrating document-level decision boundaries remain essential directions for future work.

\subsection{Robustness to AI-generated text rewriting}
\label{sec:rewriting_robustness}

Post-generation text rewriting is evaluated on matched Vietnamese passages extracted from the dataset\footnote{\url{https://doi.org/10.57967/hf/10553}} used in previous VietBinoculars research~\cite{nguyen2025vietbinoculars}. The evaluation comprises 350 eligible source passages, each exceeding 64 PhoGPT tokens after preprocessing, across three generator models including Gemma-3-12B-it, Sailor2-8B-Chat, and Sailor2-14B-Chat. For each source passage, the dataset pairs the human text with four AI variants consisting of the unmodified generation, single-step paraphrasing, two-step paraphrasing, and a prompt-guided human-style rewrite, yielding 4,550 evaluated documents in total. Each model and condition was evaluated across 350 matched Human/AI pairs using the fixed $(650,150)$ sliding-window configuration and the Youden threshold ($t^*\approx0.928$) established in Section~\ref{sec:chunking_evaluation}, without parameter retuning.

Following the document-level scoring definitions in Section~\ref{sec:chunking_evaluation}, AUROC is evaluated using negative mean chunk scores, while macro-F1 and conditional attack success rate (ASR) follow majority voting at the fixed threshold. Specifically, conditional ASR measures the proportion of AI documents correctly detected in their original form that subsequently evade detection after rewriting:
\begingroup
\[
\mathrm{ASR}_{\mathrm{cond}} =
\frac{\sum_{j=1}^{n_{\mathrm{AI}}} \mathbb{I}\{P_\text{AI}(S_j^{\mathrm{orig}})>50\%\}\,\mathbb{I}\{P_\text{AI}(S_j^{\mathrm{rewrite}})\leq50\%\}}
{\sum_{j=1}^{n_{\mathrm{AI}}} \mathbb{I}\{P_\text{AI}(S_j^{\mathrm{orig}})>50\%\}}.
\]
\endgroup
Here, $j$ indexes matched AI documents, $n_{\mathrm{AI}}$ is their number, and $S_j^{\mathrm{orig}}$ and $S_j^{\mathrm{rewrite}}$ denote the paired original and rewritten texts. For this binary evaluation, $P_\text{AI}\leq50\%$ is treated as a Human prediction. Table~\ref{tab:rewriting_robustness} reports these metrics with 95\% source-cluster bootstrap confidence intervals.

\begin{table}[!ht]
\centering
\begingroup
\caption{Paired rewriting evaluation with 350 matched Human/AI pairs per model and condition at the fixed Youden threshold ($t^*\approx0.928$) with $(W,O)=(650,150)$. AUROC is evaluated using negative mean chunk scores, while macro-F1 and conditional ASR follow majority voting. The final column reports the evasion count over originally detected AI documents, and brackets indicate 95\% source-cluster bootstrap confidence intervals. N/A denotes not applicable for the unperturbed baseline.}
\label{tab:rewriting_robustness}
\resizebox{\textwidth}{!}{%
\begin{tabular}{llcccc}
\toprule
\multirow{2}{*}{Model} & \multirow{2}{*}{Condition} & Mean-score AUROC & Majority macro-F1 & Conditional ASR & \multirow{2}{*}{ASR count} \\
& & [95\% CI] & [95\% CI] & [95\% CI] & \\
\midrule
Gemma-3-12B & Original & 0.9999 [0.9997, 1.0000] & 0.9814 [0.9714, 0.9914] & N/A & N/A \\
Gemma-3-12B & Paraphrase 1 & 0.9983 [0.9969, 0.9993] & 0.9729 [0.9600, 0.9843] & 0.0171 [0.0057, 0.0314] & 6/350 \\
Gemma-3-12B & Paraphrase 2 & 0.9984 [0.9960, 1.0000] & 0.9786 [0.9671, 0.9886] & 0.0057 [0.0000, 0.0143] & 2/350 \\
Gemma-3-12B & Human-style AI rewrite & 0.9914 [0.9840, 0.9969] & 0.9614 [0.9457, 0.9743] & 0.0400 [0.0200, 0.0629] & 14/350 \\
Sailor2-8B & Original & 1.0000 [1.0000, 1.0000] & 0.9814 [0.9714, 0.9914] & N/A & N/A \\
Sailor2-8B & Paraphrase 1 & 0.8942 [0.8661, 0.9217] & 0.8703 [0.8464, 0.8950] & 0.2200 [0.1771, 0.2629] & 77/350 \\
Sailor2-8B & Paraphrase 2 & 0.8129 [0.7763, 0.8466] & 0.7946 [0.7640, 0.8216] & 0.3629 [0.3143, 0.4143] & 127/350 \\
Sailor2-8B & Human-style AI rewrite & 0.8896 [0.8589, 0.9169] & 0.8659 [0.8404, 0.8906] & 0.2286 [0.1857, 0.2743] & 80/350 \\
Sailor2-14B & Original & 0.9459 [0.9300, 0.9617] & 0.8436 [0.8165, 0.8696] & N/A & N/A \\
Sailor2-14B & Paraphrase 1 & 0.8950 [0.8678, 0.9207] & 0.8511 [0.8253, 0.8762] & 0.2353 [0.1858, 0.2903] & 60/255 \\
Sailor2-14B & Paraphrase 2 & 0.8950 [0.8675, 0.9208] & 0.8269 [0.7982, 0.8539] & 0.2902 [0.2348, 0.3494] & 74/255 \\
Sailor2-14B & Human-style AI rewrite & 0.9031 [0.8770, 0.9272] & 0.8600 [0.8336, 0.8836] & 0.1961 [0.1506, 0.2469] & 50/255 \\
\bottomrule
\end{tabular}}
\endgroup
\end{table}

Detection robustness varied markedly across generator models and rewrite conditions. VietAIDetector demonstrated the highest resilience on Gemma-3-12B, where mean-score AUROC remained at least 0.9914 and majority macro-F1 reached at least 0.9614 across all conditions, resulting in a maximum conditional ASR of only 0.0400 (14 of 350 documents). Conversely, the detector exhibited the highest vulnerability on Sailor2-8B under Paraphrase 2, where AUROC dropped to 0.8129 and macro-F1 declined to 0.7946, yielding a conditional ASR of 0.3629 with 127 of 350 documents evading detection. For Sailor2-14B, conditional evasion ranged from 0.1961 to 0.2902, evaluated across the 255 documents correctly identified in the original baseline. The human-style AI rewrite condition reflects prompt-guided machine text rather than manual human editing. In addition, semantic preservation was not manually audited, and source-level overlap with the threshold-calibration data cannot be ruled out. Consequently, these findings represent a targeted stress test on a previously examined literary corpus rather than an evaluation of general adversarial robustness.

\subsection{OCR quality and detection performance}
\label{sec:ocr_evaluation}

This subsection assesses OCR transcription quality and downstream detection robustness in a controlled setting using 454 documents from the held-out Gemma-3-12B-News-Concat-Test benchmark. These documents constitute 227 matched pairs of human-written and AI-generated texts, selected sequentially in their fixed benchmark execution order rather than through random resampling. Each source text was rendered as an image-only clean-scan PDF and transcribed via the integrated Vintern-1B-v2 model to measure character error rate (CER) and word error rate (WER) against the ground truth. Subsequently, both the original and OCR-transcribed texts were rescored across the four chunking configurations established in Section~\ref{sec:chunking_evaluation} using the fixed Youden threshold, without retuning.

\begin{table}[!ht]
\centering
\begingroup
\caption{OCR transcription quality on the controlled clean-scan subset of Gemma-3-12B-News-Concat-Test. Micro-CER and whitespace-token WER are reported with 95\% stratified source-pair bootstrap confidence intervals from 2,000 resamples. The last two columns report pages reaching the maximum generation limit and unrecoverable page failures.}
\label{tab:ocr_quality}
\resizebox{\textwidth}{!}{%
\begin{tabular}{l rr cc rr}
\toprule
\multirow{2}{*}{Class} & \multirow{2}{*}{Documents} & \multirow{2}{*}{Pages} & Micro-CER & Whitespace-token WER & At generation limit & \multirow{2}{*}{Failed pages} \\
& & & [95\% CI] & [95\% CI] & (pages) & \\
\midrule
Human & 227 & 1,015 & 0.6358 [0.5805, 0.6932] & 0.7334 [0.6716, 0.7991] & 232 & 0 \\
AI    & 227 & 1,574 & 0.2944 [0.2565, 0.3339] & 0.3312 [0.2887, 0.3749] &  94 & 1 \\
All   & 454 & 2,589 & 0.4528 [0.4193, 0.4879] & 0.5187 [0.4808, 0.5570] & 326 & 1 \\
\bottomrule
\end{tabular}}
\endgroup
\end{table}

As reported in Table~\ref{tab:ocr_quality}, transcription error rates differed substantially between the Human and AI classes. This discrepancy is primarily driven by token truncation, as OCR generation reached the maximum output limit on 22.9\% of Human pages (232 of 1,015) compared with only 6.0\% of AI pages (94 of 1,574), causing severe deletion errors on text-dense pages. Across all 2,589 evaluated pages, 326 reached this generation ceiling, while only a single page encountered an unrecoverable extraction failure. These transcription errors, particularly the asymmetric omission of text in human-authored documents, represent an important operational constraint that directly influences downstream classification metrics.

\begin{table}[!ht]
\centering
\begingroup
\caption{Paired detection results for original text and clean-scan OCR text at the fixed Youden threshold. The evaluation includes 454 documents in 227 source pairs and uses the four configurations selected before Test evaluation. AUROC scores follow the document-level definitions in Section~\ref{sec:chunking_evaluation}. For single-score baselines, majority AUROC is based on binary decisions. Brackets report pointwise 95\% source-pair bootstrap confidence intervals from 2,000 resamples for the paired AUROC decrease (original minus clean scan) and the macro-F1 change (clean scan minus original).}
\label{tab:ocr_impact}
\resizebox{\textwidth}{!}{%
\begin{tabular}{ll ccc ccc}
\toprule
\multirow{2}{*}{Condition} & \multirow{2}{*}{Aggregation} & \multicolumn{3}{c}{AUROC} & \multicolumn{3}{c}{Macro-F1} \\
\cmidrule(lr){3-5} \cmidrule(lr){6-8}
& & Original & Clean scan & Decrease [95\% CI] & Original & Clean scan & Change [95\% CI] \\
\midrule
Full document & Majority & 0.7907 & 0.7841 & 0.0066 [$-$0.0352, 0.0463] & 0.7812 & 0.7736 & $-$0.0076 [$-$0.0539, 0.0405] \\
Full document & Mean     & 1.0000 & 0.9847 & 0.0153 [0.0041, 0.0298]   & 0.7812 & 0.7736 & $-$0.0076 [$-$0.0539, 0.0405] \\
Truncated at 768 & Majority & 0.9604 & 0.9405 & 0.0198 [$-$0.0044, 0.0441] & 0.9603 & 0.9403 & $-$0.0200 [$-$0.0444, 0.0044] \\
Truncated at 768 & Mean     & 0.9994 & 0.9789 & 0.0205 [0.0065, 0.0380]   & 0.9603 & 0.9403 & $-$0.0200 [$-$0.0444, 0.0044] \\
$(600,0)$ & Majority & 1.0000 & 0.9966 & 0.0034 [0.0004, 0.0090] & 1.0000 & 0.8632 & $-$0.1368 [$-$0.1674, $-$0.1069] \\
$(600,0)$ & Mean     & 1.0000 & 0.9023 & 0.0977 [0.0682, 0.1275] & 0.9956 & 0.7608 & $-$0.2348 [$-$0.2722, $-$0.1967] \\
$(650,150)$ & Majority & 1.0000 & 0.9951 & 0.0049 [0.0002, 0.0119] & 0.9978 & 0.8724 & $-$0.1254 [$-$0.1556, $-$0.0956] \\
$(650,150)$ & Mean     & 1.0000 & 0.8984 & 0.1016 [0.0719, 0.1320] & 0.9956 & 0.7531 & $-$0.2425 [$-$0.2825, $-$0.2044] \\
\bottomrule
\end{tabular}}
\endgroup
\end{table}

The paired results in Table~\ref{tab:ocr_impact} show clear differences across conditions. Under majority voting, sliding-window chunking yields the smallest AUROC decrease, with the primary $(650,150)$ configuration decreasing by only $0.0049$ to retain an AUROC of $0.9951$ on clean scans. In contrast, mean aggregation causes the largest performance drop, where $(650,150)$ exhibits the largest macro-F1 change of $-$0.2425 alongside an AUROC decrease of $0.1016$. While full-document scoring exhibits the smallest paired macro-F1 change ($-$0.0076), this reflects its already low original performance ($0.7812$).

\begin{table}[!ht]
\centering
\begingroup
\caption{False-positive and false-negative counts for original and clean-scan OCR text at the fixed Youden threshold across the 227 matched pairs.}
\label{tab:ocr_confusion}
\resizebox{\textwidth}{!}{%
\begin{tabular}{ll cccc}
\toprule
\multirow{2}{*}{Condition} & \multirow{2}{*}{Aggregation} & \multicolumn{2}{c}{False Positives (FP)} & \multicolumn{2}{c}{False Negatives (FN)} \\
\cmidrule(lr){3-4} \cmidrule(lr){5-6}
& & Original & Clean scan & Original & Clean scan \\
\midrule
Full document & Majority & 95 & 98 & 0 & 0 \\
Full document & Mean & 95 & 98 & 0 & 0 \\
Truncated at 768 & Majority & 17 & 27 & 1 & 0 \\
Truncated at 768 & Mean & 17 & 27 & 1 & 0 \\
$(600,0)$ & Majority & 0 & 61 & 0 & 0 \\
$(600,0)$ & Mean & 2 & 103 & 0 & 0 \\
$(650,150)$ & Majority & 1 & 57 & 0 & 0 \\
$(650,150)$ & Mean & 2 & 106 & 0 & 0 \\
\bottomrule
\end{tabular}}
\endgroup
\end{table}

Classification error counts in Table~\ref{tab:ocr_confusion} show that the drop in macro-F1 stems entirely from an asymmetric rise in false positives. On clean scans, false negatives remain at zero across all evaluated configurations, and sliding-window chunking preserves zero false negatives across both text states. These errors suggest that OCR transcription noise and text truncation alter token likelihoods in human writing, shifting scores across the fixed threshold. Consequently, while the detector maintains full recall on AI-generated content within this clean-scan evaluation subset, practical deployment on scanned documents requires threshold recalibration to mitigate false alarms. Because the evaluated PDFs are controlled clean scans, these findings characterize baseline OCR behavior rather than camera-captured or physically degraded documents.

\section{Impact}

VietAIDetector addresses the need for an open-source, zero-shot tool for detecting Vietnamese AI-generated text in practical settings. To the best of our knowledge, among the tools compared in Table~\ref{tab:performance}, it is the first open-source application specifically designed for Vietnamese that integrates zero-shot detection with an end-to-end document processing pipeline. Unlike existing open-source methods such as Binoculars~\cite{hans2024binoculars}, DetectGPT~\cite{mitchell2023detectgpt}, and GLTR~\cite{gehrmann-etal-2019-gltr}, VietAIDetector provides multi-format file processing, an interactive user interface, and downloadable PDF reports. Compared with commercial tools, it is fully open source and freely available. At the time of writing, Turnitin does not provide support for Vietnamese AI-generated text detection. By integrating the VietBinoculars algorithm~\cite{nguyen2025vietbinoculars} into a complete software package, VietAIDetector lowers the barrier to deploying AI-generated text detection for Vietnamese in research and educational settings.

VietAIDetector also supports several empirical research directions in Vietnamese AI-generated text detection. Its configurable sliding-window mechanism provides an operational framework for examining document-level aggregation and mitigating threshold drift on long texts. Furthermore, the embedded OCR pipeline enables empirical evaluation on scanned documents, establishing a baseline for transcription discrepancies and threshold recalibration. The software also facilitates stress testing against post-generation text modifications, including automated paraphrasing and prompt-guided rewriting. Finally, the modular, language-agnostic architecture facilitates adaptation of the zero-shot detection framework to other low-resource languages.

In addition to methodological contributions, VietAIDetector provides practical support for reproducible research. Embedding configuration parameters and thresholds directly within generated PDF reports allows evaluators to reproduce and verify detection results. Furthermore, structured JSON output facilitates the integration of VietAIDetector into downstream NLP pipelines and automated workflows. Together, these software features extend zero-shot AI-generated text detection research~\cite{nguyen2025vietbinoculars,hans2024binoculars} to practical Vietnamese document analysis in research and educational workflows.

Practical deployment of the software spans two primary user groups. In \textit{higher education}, educators can screen student assignments, essays, and theses for potential AI misuse through a web interface and generate structured PDF reports for documentation and review. This is particularly relevant in Vietnam, where AI-assisted academic dishonesty has become an increasing concern. Crucially, detection results are intended to serve as supporting evidence for pedagogical dialogue and holistic review, ensuring that academic integrity decisions are never made autonomously without comprehensive human oversight. 
In \textit{social media content verification}, fact-checkers, content reviewers, and the general public can analyze suspected articles or posts by pasting raw text or uploading documents, while chunk-level results help identify sections that are likely AI-generated. 
In addition, the provided Kaggle deployment script (\texttt{run\_kaggle.sh}) enables deployment on free-tier dual NVIDIA T4 GPUs, making the software readily accessible to individual users and small organizations without requiring dedicated computing infrastructure.

VietAIDetector is released under the MIT License and is freely available on GitHub. Unlike commercial tools such as GPTZero~\cite{adam2026gptzeroro} and Turnitin, it is fully open source, transparent, and readily extensible for community contributions and institutional customization. The software is planned for pilot deployment at Nha Trang University to support academic integrity assessment. Its zero-shot design also enables adaptation to newly emerging Vietnamese large language models without retraining.

\section{Limitations and future work}

Despite its advantages, VietAIDetector has several limitations. Detection performance depends on the quality of the underlying language models and may vary across domains. Although the sliding-window chunking mechanism enables long-document processing, selecting optimal chunking parameters remains an open research problem. The exploratory benchmark uses only 20 documents per dataset and reuses these datasets for configuration selection and evaluation, which may yield optimistic performance estimates. Furthermore, the large-scale benchmark relies on concatenated news passages, which may not capture the discourse coherence of naturally authored long articles. Detection in OCR-degraded documents and documents containing complex structures, such as tables, images, or multimedia content, also remains challenging. The OCR assessment uses controlled clean scans from a fixed benchmark subset and does not establish performance on camera-captured or physically degraded documents. Similarly, the rewriting evaluation represents a bounded stress test across automated generators and does not evaluate human-in-the-loop editing or adaptive evasion attacks. In addition, improving detection accuracy often requires larger language models, resulting in higher computational costs. As LLMs continue to evolve, detection methods will require continuous adaptation to maintain their effectiveness~\cite{varshney2020limits}. Furthermore, the tool has not yet been extensively evaluated in diverse real-world deployment scenarios.
VietAIDetector is designed to assist in assessing the likelihood of AI-generated content, providing probabilistic supporting evidence rather than definitive proof of AI authorship. Consequently, the tool should not be regarded as an autonomous or legally authoritative decision-making system. Final decisions regarding document authenticity and academic integrity must always be made by human evaluators.

Future work will focus on evaluating naturally authored long documents, investigating threshold recalibration under severe physical OCR degradation, and improving resilience against human-edited and adaptive rewriting attacks. In addition, extending the software to support complex layouts containing tables and embedded formulas remains an important engineering objective. The planned pilot deployment at Nha Trang University will also provide real-world data for evaluating the tool across diverse educational scenarios and guiding further improvements in accuracy and adaptability.

\section{Conclusions}

This paper presents VietAIDetector, an open-source tool for detecting Vietnamese AI-generated text based on the VietBinoculars and Binoculars frameworks~\cite{nguyen2025vietbinoculars,hans2024binoculars}. By adopting a zero-shot approach, the tool detects AI-generated content without model retraining. VietAIDetector supports multiple input formats, including scanned PDFs and long documents via sliding-window chunking, providing chunk-level inspection together with configurable thresholds and downloadable PDF reports. Empirical evaluations indicate that the software mitigates threshold drift on long texts and enables end-to-end processing of scanned documents. Under automated paraphrasing, detection resilience varies markedly across generator models, maintaining high robustness on Gemma-3-12B while exhibiting increased vulnerability on Sailor2.

The software is built on a modular architecture with a configurable processing pipeline, facilitating maintenance, future extensions, and integration into downstream NLP applications through structured JSON output. Released under the MIT License and publicly available on GitHub, VietAIDetector provides an open, transparent, and practical platform for Vietnamese AI-generated text detection in both research and real-world applications.

\section*{Acknowledgements}
This research was funded by Nha Trang University under grant number TR2026-13-07. The authors thank the open-source community for developing the tools, libraries, and LLMs that made this software possible. We also extend our thanks to the reviewers for their helpful comments and suggestions.
\bibliographystyle{IEEEtranN}
\bibliography{references}

\newpage
\appendix
\renewcommand{\thesection}{Appendix \Alph{section}}

\section{Calibration dataset and updated detection thresholds}\label{appendix:update-thresholds}
\begin{figure}[!htb]

    \centering

    \includegraphics[width=1.0\textwidth]{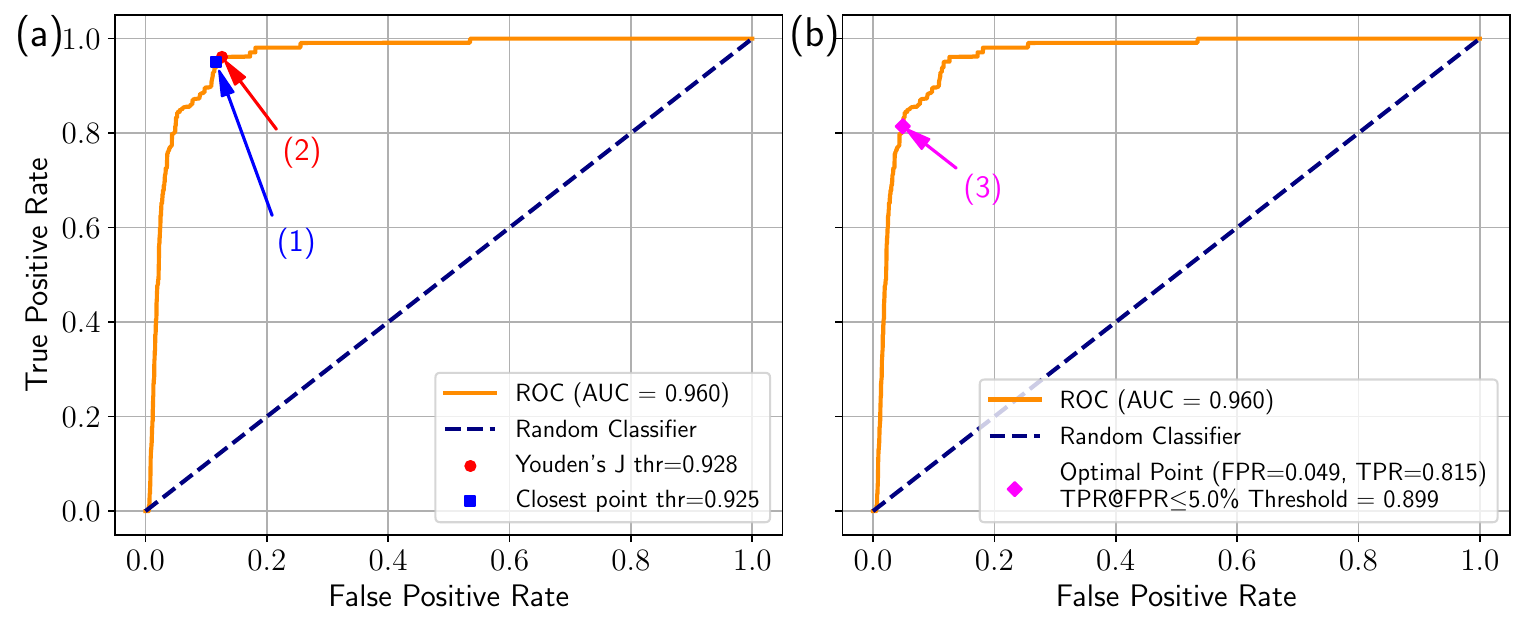}

    \caption{Detection thresholds updated in July 2026 using VietAIDetector-Calibration-News-2026-07, which contains 10,000 Human and 10,000 AI records. (a) Youden and closest-point thresholds. (b) TPR@5\%FPR threshold.}

    \label{fig:updated_thresholds}

\end{figure}

As discussed above, keeping pace with increasingly sophisticated LLMs that generate human-like text requires periodically updating the decision thresholds of VietAIDetector. For the July 2026 update, the calibration dataset, designated as VietAIDetector-Calibration-News-2026-07 and preserved under \texttt{benchmarks/train\_datasets}, contains 20,000 records equally partitioned between Human and AI classes. The AI partition comprises 5,000 articles from Google Gemini-3.7-Flash and 5,000 articles from OpenAI GPT-5.6-Luna, generated through topical journalistic prompts across six domains spanning politics, economics, society, culture, sports, and education. Human texts originate from our previous benchmark~\cite{nguyen2025vietbinoculars} and are sampled via length stratification to match the empirical token distribution of machine text (Figure~\ref{fig:token_length_distribution}a). This stratification substantially mitigates sequence length disparities across classes, although it does not entirely rule out residual confounding. Crucially, this collection is dedicated exclusively to threshold calibration and remains strictly decoupled from the chunking development and held-out test evaluation sets. Calibration procedures determine three operational thresholds from the empirical ROC curve, which achieves an AUROC of 0.960. These thresholds comprise the Youden index ($J = \mathrm{TPR} - \mathrm{FPR}$, yielding $t^* \approx 0.928$), the closest point to $(0, 1)$ ($t^* \approx 0.925$), and the low-FPR criterion maximizing TPR under $\mathrm{FPR} \le 5\%$ ($t^* \approx 0.899$). Figure~\ref{fig:updated_thresholds} illustrates these calibrated operating points, which users can integrate directly into VietAIDetector by modifying \texttt{settings.py}.

\section{Exploratory survey of chunking configurations}\label{appendix:grid_search}
\begingroup
\scriptsize
\setlength{\tabcolsep}{1.5pt}
\setlength{\LTcapwidth}{\textwidth}
\renewcommand{\arraystretch}{1.15}
\begin{longtable}{@{}*{12}{>{\centering\arraybackslash}p{\dimexpr\textwidth/12-2\tabcolsep\relax}}@{}}
\caption{Exploratory results on Claude-Sonnet-4.6-News, Gemini-3.7-Flash-News, and GPT-5.6-Luna-News across chunking parameters ($W \in [200, 650]$, $O \in [50, 150]$). The Youden threshold is $t^{*} = 0.927966$, with $N=20$ per dataset.}
\label{tab:grid_summary_compact}\\
\toprule
\multicolumn{2}{c}{\textbf{Chunking}} & \multicolumn{3}{c}{\textbf{Claude-Sonnet-4.6}} & \multicolumn{3}{c}{\textbf{Gemini-3.7-Flash}} & \multicolumn{3}{c}{\textbf{GPT-5.6-Luna}} & \textbf{Time} \\
\cmidrule(lr){1-2} \cmidrule(lr){3-5} \cmidrule(lr){6-8} \cmidrule(lr){9-11} \cmidrule(lr){12-12}
 $W$ & $O$ & \textbf{AI\%} & \textbf{Chunks} & \textbf{Recall} & \textbf{AI\%} & \textbf{Chunks} & \textbf{Recall} & \textbf{AI\%} & \textbf{Chunks} & \textbf{Recall} & \textbf{(s)} \\
\midrule
\endfirsthead
\caption[]{Exploratory survey of chunking configurations (continued).}\\
\toprule
\multicolumn{2}{c}{\textbf{Chunking}} & \multicolumn{3}{c}{\textbf{Claude-Sonnet-4.6}} & \multicolumn{3}{c}{\textbf{Gemini-3.7-Flash}} & \multicolumn{3}{c}{\textbf{GPT-5.6-Luna}} & \textbf{Time} \\
\cmidrule(lr){1-2} \cmidrule(lr){3-5} \cmidrule(lr){6-8} \cmidrule(lr){9-11} \cmidrule(lr){12-12}
 $W$ & $O$ & \textbf{AI\%} & \textbf{Chunks} & \textbf{Recall} & \textbf{AI\%} & \textbf{Chunks} & \textbf{Recall} & \textbf{AI\%} & \textbf{Chunks} & \textbf{Recall} & \textbf{(s)} \\
\midrule
\endhead
\midrule
\multicolumn{12}{r}{\textit{Continued on next page.}}\\
\endfoot
\bottomrule
\endlastfoot
200 & 50 & 81.98 & 14.95 & 90.0 & 59.44 & 15.10 & 65.0 & 67.81 & 16.00 & 90.0 & 0.41 \\
200 & 100 & 82.24 & 21.60 & 95.0 & 55.62 & 22.05 & 55.0 & 69.13 & 23.00 & 95.0 & 0.54 \\
200 & 150 & 82.73 & 41.80 & 100.0 & 57.11 & 42.60 & 65.0 & 69.34 & 44.40 & 95.0 & 1.05 \\
\addlinespace[3pt]
250 & 50 & 80.99 & 11.25 & 90.0 & 63.37 & 11.55 & 75.0 & 73.33 & 12.00 & 90.0 & 0.36 \\
250 & 100 & 86.38 & 14.60 & 100.0 & 59.10 & 14.95 & 65.0 & 70.52 & 15.40 & 95.0 & 0.45 \\
250 & 150 & 84.51 & 21.20 & 95.0 & 60.41 & 21.55 & 70.0 & 74.27 & 22.40 & 100.0 & 0.66 \\
\addlinespace[3pt]
300 & 50 & 90.00 & 9.00 & 100.0 & 66.89 & 9.10 & 75.0 & 78.00 & 10.00 & 100.0 & 0.38 \\
300 & 100 & 87.73 & 11.00 & 95.0 & 66.59 & 11.10 & 85.0 & 77.50 & 12.00 & 100.0 & 0.45 \\
300 & 150 & 90.31 & 14.25 & 100.0 & 65.52 & 14.55 & 75.0 & 80.33 & 15.00 & 95.0 & 0.55 \\
\addlinespace[3pt]
350 & 50 & 88.12 & 7.95 & 95.0 & 66.88 & 8.00 & 70.0 & 76.25 & 8.00 & 80.0 & 0.34 \\
350 & 100 & 87.22 & 9.00 & 95.0 & 66.67 & 9.00 & 80.0 & 74.56 & 9.40 & 85.0 & 0.41 \\
350 & 150 & 87.73 & 10.95 & 95.0 & 65.91 & 11.00 & 80.0 & 79.47 & 11.40 & 95.0 & 0.50 \\
\addlinespace[3pt]
400 & 50 & 87.14 & 6.95 & 100.0 & 70.71 & 7.00 & 75.0 & 88.57 & 7.00 & 100.0 & 0.35 \\
400 & 100 & 92.95 & 7.60 & 95.0 & 69.82 & 7.95 & 75.0 & 90.00 & 8.00 & 100.0 & 0.38 \\
400 & 150 & 90.56 & 8.95 & 100.0 & 65.56 & 9.00 & 75.0 & 92.22 & 9.00 & 100.0 & 0.45 \\
\addlinespace[3pt]
450 & 50 & 88.33 & 6.00 & 95.0 & 69.17 & 6.00 & 80.0 & 92.50 & 6.00 & 100.0 & 0.36 \\
450 & 100 & 89.64 & 6.60 & 95.0 & 67.74 & 6.95 & 70.0 & 92.86 & 7.00 & 100.0 & 0.39 \\
450 & 150 & 94.11 & 7.25 & 95.0 & 68.21 & 7.55 & 80.0 & 92.50 & 8.00 & 100.0 & 0.43 \\
\addlinespace[3pt]
500 & 50 & 96.00 & 5.00 & 100.0 & 78.17 & 5.10 & 85.0 & 90.00 & 6.00 & 100.0 & 0.36 \\
500 & 100 & 91.67 & 6.00 & 95.0 & 76.67 & 6.00 & 80.0 & 95.00 & 6.00 & 100.0 & 0.39 \\
500 & 150 & 94.52 & 6.25 & 100.0 & 74.64 & 6.55 & 80.0 & 95.71 & 7.00 & 100.0 & 0.43 \\
\addlinespace[3pt]
550 & 50 & 88.00 & 5.00 & 100.0 & 74.00 & 5.00 & 85.0 & 97.00 & 5.00 & 100.0 & 0.38 \\
550 & 100 & 94.00 & 5.00 & 95.0 & 70.00 & 5.00 & 85.0 & 94.83 & 5.40 & 100.0 & 0.39 \\
550 & 150 & 90.00 & 5.95 & 95.0 & 71.67 & 6.00 & 80.0 & 96.67 & 6.00 & 100.0 & 0.44 \\
\addlinespace[3pt]
600 & 50 & 91.75 & 4.25 & 95.0 & 75.50 & 4.55 & 80.0 & 92.00 & 5.00 & 100.0 & 0.39 \\
600 & 100 & 90.00 & 5.00 & 95.0 & 69.00 & 5.00 & 75.0 & 96.00 & 5.00 & 100.0 & 0.41 \\
600 & 150 & 96.00 & 5.00 & 100.0 & 81.00 & 5.00 & 90.0 & 100.00 & 5.00 & 100.0 & 0.41 \\
\addlinespace[3pt]
650 & 50 & 95.00 & 4.00 & 95.0 & 78.75 & 4.00 & 80.0 & 97.50 & 4.00 & 100.0 & 0.39 \\
650 & 100 & 95.00 & 4.00 & 100.0 & 79.25 & 4.10 & 85.0 & 92.00 & 5.00 & 100.0 & 0.41 \\
650 & 150 & 90.00 & 4.95 & 100.0 & 74.00 & 5.00 & 85.0 & 95.00 & 5.00 & 100.0 & 0.46 \\
\end{longtable}
\endgroup
$W$ and $O$ denote window size and overlap in tokens, respectively. Column groups correspond to the three evaluated news datasets, specifically Claude-Sonnet-4.6-News, Gemini-3.7-Flash-News, and GPT-5.6-Luna-News. AI\% indicates the mean percentage of AI-classified chunks per document, Chunks indicates the mean chunk count per document, and Recall denotes the document-level detection rate on the 20 AI-generated documents under majority voting. Time represents the average processing time per document in seconds.

\section{Development-set selection of chunking configurations}
\label{appendix:development_selection}

Fifteen overlapping and three non-overlapping configurations achieved both macro-F1 and AUROC of 1.0000 on the development set. Within these tied groups, $(650,150)$ and $(600,0)$ minimized the mean number of chunks per document, averaging 5.287 and 4.660, respectively. These configurations were retained for test evaluation.

\begin{figure}[!htb]
\centering
\includegraphics[width=0.88\textwidth]{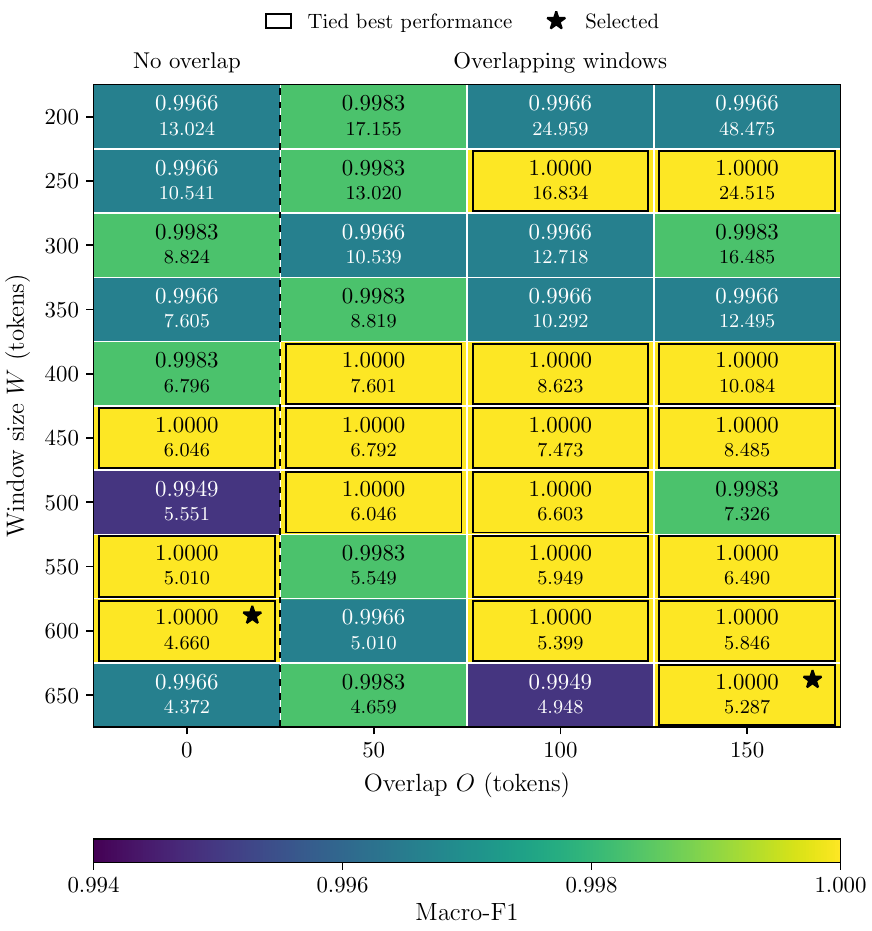}
\caption{Development results for 40 configurations on 592 Gemma-3-12B-News-Concat-Dev documents under majority voting at the fixed Youden threshold. Each cell reports macro-F1 above mean chunks per document. Color represents macro-F1 on a restricted scale from 0.994 to 1.000. Outlined cells achieve both macro-F1 and AUROC of 1.0000. Stars identify the fewest-chunk configurations among tied candidates, selected separately for overlapping and non-overlapping windows.}
\label{fig:development_chunking_selection}
\end{figure}

\section{Token-length distributions of calibration and benchmark datasets}
\label{appendix:token_length_distributions}

\begin{figure}[!htb]
\centering
\includegraphics[width=0.92\textwidth]{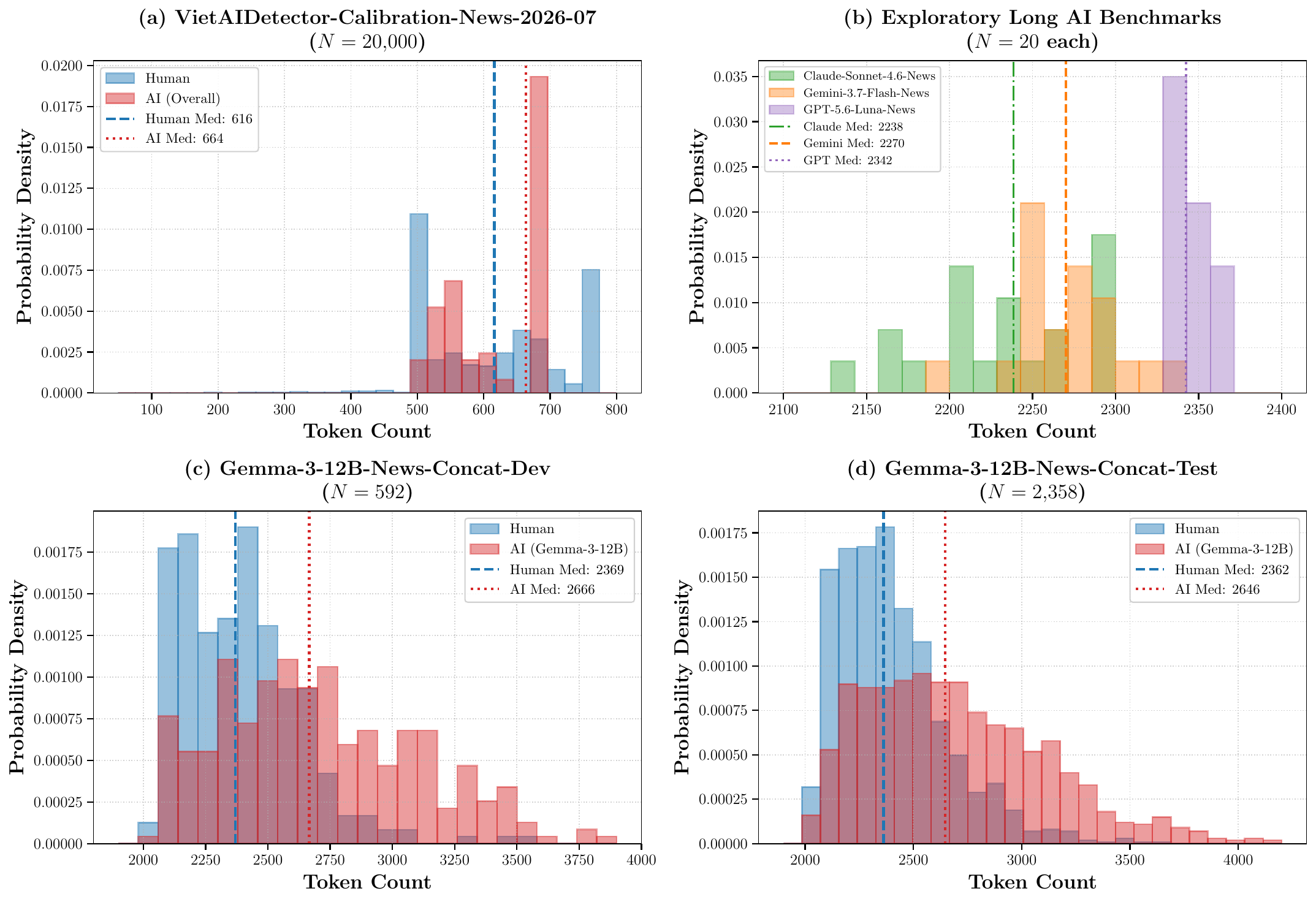}
\caption{Empirical token-length probability distributions across evaluation and calibration datasets. (a) VietAIDetector-Calibration-News-2026-07 with 10,000 Human and 10,000 AI records. (b) Exploratory long AI benchmarks with 20 documents each for Claude-Sonnet-4.6-News, Gemini-3.7-Flash-News, and GPT-5.6-Luna-News. (c) Gemma-3-12B-News-Concat-Dev with 296 Human and 296 AI documents. (d) Gemma-3-12B-News-Concat-Test with 1,179 Human and 1,179 AI documents. Vertical dashed and dotted lines indicate median token counts for Human and AI classes, respectively.}
\label{fig:token_length_distribution}
\end{figure}

Figure~\ref{fig:token_length_distribution} presents the empirical token-length probability distributions across all calibration and evaluation datasets. While calibration and exploratory sets span distinct length regimes in Figure~\ref{fig:token_length_distribution}(a)--(b), concatenated documents in Figure~\ref{fig:token_length_distribution}(c)--(d) exhibit substantial distributional overlap between human and AI texts across 2,000 to 4,000 tokens. This broad overlap shows that document concatenation substantially mitigates length disparities between classes. However, because modest distributional differences remain, this design does not entirely rule out sequence length as a potential confounding factor.

\end{document}